\pdfoutput=1

\documentclass[11pt]{article}

\usepackage{acl}

\usepackage{times}
\usepackage{latexsym}

\usepackage[T1]{fontenc}

\usepackage[utf8]{inputenc}

\usepackage{microtype}
\usepackage{tabularx,ragged2e}
\usepackage{hyperref}
\usepackage{setspace}
\usepackage{caption}
\usepackage{multirow}
\usepackage{subcaption}
\usepackage{graphicx}
\usepackage{makecell}
\usepackage{pgfplots}
\usepackage{todonotes}
\usepackage{hyperref}
\usepackage{amsmath}
\usepackage{pifont}
\usepackage{subcaption,booktabs}

\title{Description-Driven Task-Oriented Dialog Modeling}

\author{Jeffrey Zhao, Raghav Gupta, Yuan Cao, Dian Yu, Mingqiu Wang, \\
\textbf{Harrison Lee, Abhinav Rastogi, Izhak Shafran, Yonghui Wu} \\
Google Research \\
\texttt{\{jeffreyzhao, raghavgupta, yuancao\}@google.com}}

\begin{document}
\maketitle
\begin{abstract}

Task-oriented dialogue (TOD) systems are required to identify key information from conversations for the completion of given tasks. Such information is conventionally specified in terms of intents and slots contained in task-specific ontology or schemata. Since these schemata are designed by system developers, the naming convention for slots and intents is not uniform across tasks, and may not convey their semantics effectively. This can lead to models memorizing arbitrary patterns in data, resulting in suboptimal performance and generalization.
In this paper, we propose that schemata should be modified by replacing names or notations entirely with natural language descriptions. We show that a language description-driven system exhibits better understanding of task specifications, higher performance on state tracking, improved data efficiency, and effective zero-shot transfer to unseen tasks. 
Following this paradigm, we present a simple yet effective \textbf{D}escription-\textbf{D}riven \textbf{D}ialog \textbf{S}tate \textbf{T}racking (D3ST) model, which relies purely on schema descriptions and an ``index-picking'' mechanism. We demonstrate the superiority in quality, data efficiency and robustness of our approach as measured on the MultiWOZ \citep{budzianowski-etal-2018-multiwoz}, SGD \cite{Rastogi_Zang_Sunkara_Gupta_Khaitan_2020}, and the recent SGD-X \citep{lee2021sgdx} benchmarks.

\end{abstract}

\section{Introduction}\label{sec:intro}
The design of a task-oriented dialog (TOD) system conventionally starts with defining a schema specifying the information required to complete its tasks --- usually, a list of relevant slots and intents. These slots and intents often appear as abbreviated notations, such as \texttt{train-leaveat} and \texttt{hotel-internet}, to indicate the domain of a task and the information it captures.

Models are trained using these schemata will be heavily dependent on these abbreviations. This is especially true for decoder-only or sequence-to-sequence (seq2seq) TOD models, which are often trained with supervision to predict dialogue belief states as sequences of these notations. For example, a sequence such as \texttt{train-leaveat=3:00pm, hotel-internet=no}, or sequences of similar structure, are the target output for TOD models described by  \citealt{hosseiniasl2020simple} and \citealt{zhao-etal-2021-effective-sequence}.

This has several disadvantages. First, the element notations convey little semantic (and possibly ambiguous) meaning for the requirements of the slot \cite{du-etal-2021-qa}, potentially harming language understanding. Second, task-specific abstract schema notations make it easy for a model to overfit on observed tasks and fail to transfer to unseen ones, even if there is sufficient semantic similarity between the two. Finally, creating notations for each slot and intent complicates the schema design process.

In this paper, we advocate for TOD schemata with intuitive, human-readable, and semantically-rich natural language descriptions, rather than the abbreviated notations that have become customary when designing TOD models. Instead of ``\texttt{hotel-internet}'', we assert that it is more natural to describe this slot as ``\texttt{whether the hotel has internet}''. This would be easier for both the designer of the TOD system when specifying the task ontology, and we also argue that it plays an important role in improving model quality and data efficiency. 

To this end, we present a simple yet effective approach: \textbf{D}escription-\textbf{D}riven \textbf{D}ialog \textbf{S}tate \textbf{T}racking (D3ST). Here, schema descriptions are indexed and concatenated as prefixes to a seq2seq model, which then learns to predict active schema element indices and corresponding values. In addition, an index-picking mechanism reduces the chance of the model overfitting to specific schema descriptions. We demonstrate its superior performance measured on benchmarks including MultiWOZ \citep{budzianowski-etal-2018-multiwoz,zang-etal-2020-multiwoz,han2021multiwoz,ye2021multiwoz} and Schema-Guided Dialogue (SGD, \citep{Rastogi_Zang_Sunkara_Gupta_Khaitan_2020}), as well as strong few- and zero-shot transfer capability to unseen tasks. We also show evidence that, under this very general setting, natural language descriptions lead to better quality over abbreviated notations.

\section{Related Work}
In recent years, there has been increasing interest in leveraging language prompts for data efficiency and quality improvement for dialogue modelling.

\textbf{Inclusion of task descriptions}: One line of research focuses on providing descriptions or instructions related to the dialogue tasks. \citet{shah-etal-2019-robust} utilized both slot descriptions and a small number of examples of slot values for learning slot representations for spoken language understanding. Similar to our work,
\citet{lin-etal-2021-leveraging,lee2021dialogue} provided slot descriptions as extra inputs to the model and have shown quality improvement as well as zero-shot transferability. \citet{mi2021cins} extended the descriptions to a more detailed format by including task instructions, constraints and prompts altogether, demonstrating advantages of providing more sophisticated instructions to the model. However, unlike our approach, they predict slot values one-by-one in turn, which becomes increasingly inefficient as the number of slots increases, and is also prone to oversampling slot values since most slots are inactive at any stage during a dialogue. In contrast, our work predicts all states in a single pass, and is hence more efficient.

\textbf{Prompting language models}: Powerful language models like GPT \citep{radford2019language,NEURIPS2020_1457c0d6} demonstrated impressive few-shot learning ability even without fine-tuning. It is therefore natural to consider leveraging these models for few-shot dialogue modeling. \citet{madotto2020language} applied GPT-2 by priming the model with examples for language understanding, state tracking, dialogue policy and language generation tasks respectively, and in \citet{madotto2021fewshot} this approach has been extended to systematically evaluate on a set of diversified tasks using GPT-3 as backbone. Unlike these works in which the language models are frozen, we finetune the models on downstream tasks. \citet{budzianowski-vulic-2019-hello,peng2020scgpt} on the other hand, applied GPT-2 for few-shot and transferable response generation with given actions, whereas our work focuses mainly on state tracking.

\textbf{Describe task with questions}: Another line of research casts state tracking as a question answering (QA) or machine reading (MR) problem \citep{gao-etal-2020-machine,namazifar2020language,li-etal-2021-zero,lin2021zeroshot}, in which models are provided questions about each slot and their values are predicted as answers to these questions. The models are often finetuned on extractive QA or MR datasets, and by converting slot prediction into QA pairs the models are able to perform zero-shot state tracking on dialogue datasets. Their question generation procedure however, is more costly than using schema descriptions, which we adopt in our work.

\section{Methodology} \label{sec:method}

D3ST uses a seq2seq model for dialogue state tracking, and relies purely on descriptions of schema items to instruct the model.

\subsection{Model}
We choose to use seq2seq for modeling for the following reasons: first, seq2seq is a general and versatile architecture that can easily handle different formats of language instructions; second, seq2seq has been shown to be an effective approach for DST \citep{zhao-etal-2021-effective-sequence}; and third, seq2seq as a generic model architecture can be easily initialized from a publicly available pretrained checkpoint.

For D3ST, we used the T5 \cite{2020t5} model and the associated pretrained checkpoints of different sizes: Base (220M parameters), Large (770M parameters), and XXL (11B parameters).

\subsection{Description-Driven Modeling} \label{subsec:d3st}
\begin{figure*}[htbp]
\includegraphics[width=16cm]{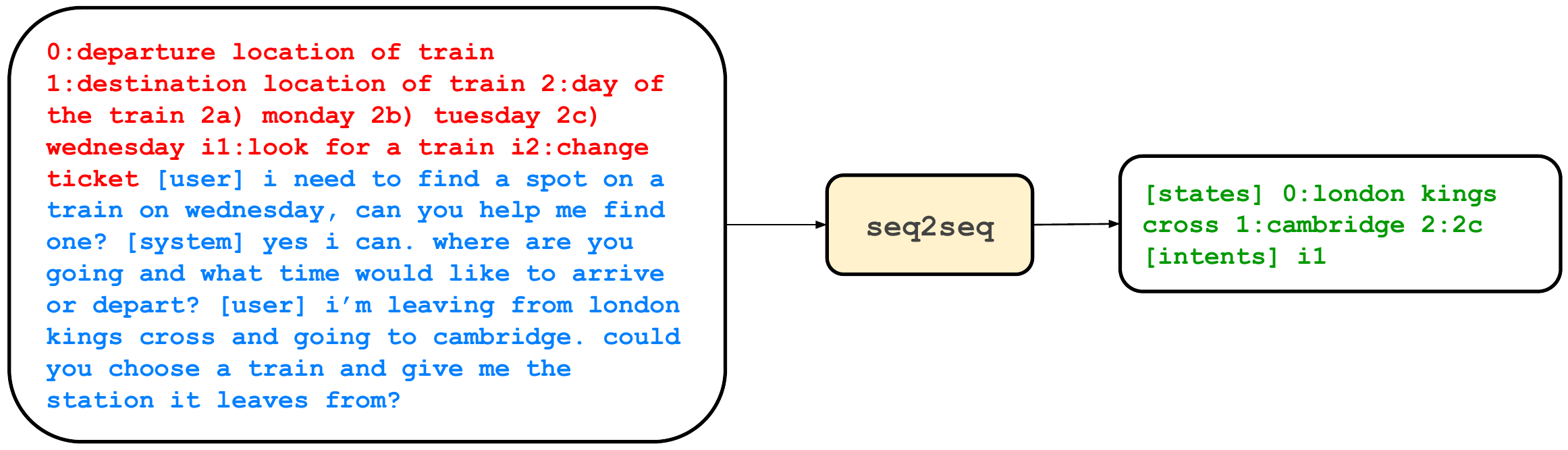}
\centering
\caption{An example of D3ST. Red: Indexed schema description sequence as prefix; Blue: Conversation context; Green: State prediction sequence. See Section \ref{sec:method} for details. Best viewed in color.}
\label{figure:d3st}
\end{figure*}

D3ST relies solely on schema descriptions for dialogue state tracking. An example of D3ST is provided in Figure \ref{figure:d3st}.

Given a set of descriptions corresponding to slots and intents specified by a schema, let $\mathtt{d_i^{slot}, i=1\ldots N}$ and $\mathtt{d_j^{intent}, j=1\ldots M}$ be the descriptions for slots and intents respectively, where $\mathtt{N}$ and $\mathtt{M}$ are the numbers of slots and intents. Let $\mathtt{u_t^{usr}}$ and $\mathtt{u_t^{sys}}$ be the user and system utterance at turn $\mathtt{t}$ respectively. 

\textbf{Input} The input to the encoder consists of the slot descriptions, intent descriptions, and conversation context concatenated into a single string. The slot descriptions have the following format:
\begin{align*}
\mathtt{0:d_0^{slot} \ldots I:d_S^{slot}}
\end{align*}
Similarly, the intent descriptions have the following format:
\begin{align*}
\mathtt{\mathbf{i}0:d_0^{int} \ldots \mathbf{i}J:d_J^{int}}
\end{align*}
\noindent Note that $\mathtt{0\ldots I}$ and $\mathtt{\mathbf{i}0 \ldots \mathbf{i}J}$ are the indices we assign to each of the slot and intent descriptions respectively. Here, ``$\mathbf{i}$'' is a literal character to differentiate intent indices from those for slots. To prevent the model from memorizing association between a specific index:description pair, we randomize the assignment of indices to descriptions for each example during training. Such a dynamic construction forces the model to consider descriptions rather than treating inputs as constant strings to make generalizable predictions. The conversation context consists of all turns of the conversations concatenated together, with leading $\mathtt{[user]}$ and $\mathtt{[sys]}$ tokens before each user and system utterance, signalling the speaker of each utterance.
\begin{align*}
\mathtt{[usr]\,u_0^{usr}\,[sys]\,u_0^{sys} \ldots [usr]\,u_T^{usr}\,[sys] \,u_T^{sys}}
\end{align*}

\textbf{Output} The decoder generates a sequence of dialogue states in the format
\begin{align*}
\mathtt{[states]\, a_0^s:v_0^s \ldots a_M^s:v_M^s\,[intents]\, a_0^i\ldots a_N^i}
\end{align*}
where $\mathtt{a_m^s}$ is the index of the $\mathtt{m}^{th}$ \emph{active} slot and there are $\mathtt{M}$ active slots in all, $\mathtt{v_m^s}$ is its corresponding value. $\mathtt{a_n^i}$ is the index of the $\mathtt{n}^{th}$ active intent and $\mathtt{N}$ is the number of active intents. This way the model learns to identify active schema elements with abstract indices, as we randomize the element order during training. Note that inactive elements are not generated.

\textbf{Handling categorical slots} Some slots are categorical, that is, they have pre-defined candidate values for the model to choose from. For example ``\texttt{whether the hotel provides free wifi or not}'' could have the categorical values ``\texttt{yes}'' and ``\texttt{no}''. To improve categorical slot prediction accuracy, we enumerate possible values together with their slot descriptions. That is, assuming the $\mathtt{i}^{th}$ slot is categorical and has $\mathtt{k}$ values $\mathtt{v_a \ldots v_k}$, its corresponding input format is
\begin{align*}
\mathtt{i:d_i^{slot} \, ia) \, v_a \ldots ik) \, v_k}
\end{align*}
in which $\mathtt{ia)}\ldots \mathtt{ik)}$ are indices assigned to each of the values.\footnote{One may also adopt $\mathtt{a)}\ldots \mathtt{k)}$ as value indices or even completely discard indexing for categorical values, however we found this shared indexing across categorical slots can sometimes cause selection ambiguity when some values (like ``\texttt{true}'' or ``\texttt{false}'') are shared by multiple categorical slots. We therefore apply slot-specific indices $\mathtt{ia)}\ldots \mathtt{ik)}$ to constrain index-picking within the $\mathtt{i}^{th}$ slot value range.} Assuming this slot is active with its third value ($\mathtt{v_c}$) being mentioned, then the corresponding prediction has the format $\mathtt{i:ic)}$.

\subsection{Properties}
From the formulation described in Section \ref{subsec:d3st}, we expect our proposed approach to have the following properties. First, the model relies fully on the understanding of schema descriptions for the identification of active slots and intents. Second, the model learns to pick indices corresponding to the active slots, intents or categorical values, instead of generating these schema elements. This ``index-picking'' mechanism, based on schema description understanding, reduces the chance of the model memorizing training schemata and makes it easier for the model to zero-shot transfer to unseen tasks. Finally, unlike previous work which also takes advantage of schema descriptions (for example \citealp{lin-etal-2021-leveraging,lee2021dialogue}) but generates values for each slot in turn (even if a slot is inactive), our approach enables predicting multiple active (and only active) slot-value pairs together with intents with a single decoding pass, making the inference procedure more efficient.

We also note that the sequence of schema descriptions prepended to the conversation context plays a similar role as instructions for specific tasks \citep{wei2021finetuned,mishra2021crosstask}. Providing more detailed human-readable descriptions enables the language model understand task requirements better, and leads to improved few-shot performance, as will be seen in experimental results.

\section{Experiments}
We design our experiments to answer the following questions:
\begin{enumerate}
    \item What is the quality of the D3ST model, when all training data is available?
    \item How does the description type for schema definition, including human-readable natural descriptions, abbreviated or even random notations, affect model quality?
    \item How data-efficient is D3ST in the low-resource or zero-shot regimes, and how do different description types affect efficiency?
    \item How robust is the model to different wordings of the human-readable descriptions?
\end{enumerate}

\subsection{Setup}

\textbf{Datasets} We conduct experiments on the MultiWOZ 2.1-2.4 \citep{budzianowski-etal-2018-multiwoz,zang-etal-2020-multiwoz,han2021multiwoz,ye2021multiwoz} and SGD \citep{Rastogi_Zang_Sunkara_Gupta_Khaitan_2020} datasets. The MultiWOZ dataset is known to contain annotation errors in multiple places and previous work adopted different data pre-processing procedures, so we follow the recommended procedure\footnote{\url{https://github.com/budzianowski/multiwoz\#dialog-state-tracking}} of using the TRADE \citep{wu-etal-2019-transferable} script to pre-process MultiWOZ 2.1. However, we do not apply any pre-processing to 2.2-2.4 for reproducibility and fair comparison with existing results. We use Joint Goal Accuracy (JGA) as the evaluation metric, which measures the percentage of turns across all conversations for which all states are correctly predicted by the model.

\textbf{Training setup} We use the open-source T5 code base\footnote{\url{https://github.com/google-research/text-to-text-transfer-transformer}} and the associated T5 1.1 checkpoints.\footnote{\url{https://github.com/google-research/text-to-text-transfer-transformer/blob/main/released\_checkpoints.md}} We consider models of the size base (250M parameters), large (800M) and XXL (11B) initialized from the corresponding pretrained checkpoints, and ran each experiment on 64 TPU v3 chips \cite{tpu2017}. For fine-tuning, we use batch size 32 and use constant learning rate of $1e-4$ across all experiments. The input and output sequence lengths are 1024 and 512 tokens, respectively.

\textbf{Descriptions} We use the slot and intent descriptions included in the original MultiWOZ and SGD datasets as inputs ($\mathtt{d_i^{slot}}$ and $\mathtt{d_i^{int}}$ described in Section \ref{subsec:d3st}) to the model. For MultiWOZ, we include schema descriptions across all domains as model prefix and set the input length limit to 2048. To avoid ambiguity between descriptions from different domains, we also add domain names as part of the descriptions. For example for the \texttt{hotel-parking} slot, the description is ``\texttt{hotel-parking facility at the hotel}''. For SGD, we include descriptions from domains relevant to each turn as suggested by the standard evaluation.

\subsection{Main Results} \label{sec:main_result}

\begin{table*}[ht!]
\begin{center}
\begin{subtable}[ht]{\textwidth}
    \centering
    \scalebox{0.85}{
    \begin{tabular}{lcccccc}
    \hline
    \textbf{Model} & \textbf{Pretrain. Model (\# Params.)} & \textbf{MW2.1} & \textbf{MW2.2} & \textbf{MW2.3} & \textbf{MW2.4} \\
    \hline
    Transformer-DST \cite{zeng2021jointly} & BERT Base (110M) & 55.35 & - & - & -\\
    SOM-DST \cite{kim2020efficient} & BERT Base (110M) & 51.2 & - & 55.5 & 66.8\\
    TripPy \cite{heck-etal-2020-trippy} & BERT Base (110M) & 55.3 & - & \textbf{63.0} & 59.6\\
    SAVN \cite{wang-etal-2020-slot} & BERT Base (110M) & 54.5 & - & 58.0 & 60.1\\
    SimpleTOD\ding{72} \cite{hosseiniasl2020simple} & DistilGPT-2 (82M) & 50.3/55.7 & - & 51.3 & -\\
    Seq2seq \cite{zhao-etal-2021-effective-sequence} & T5 Base (220M) & 52.8 &57.6 & 59.3 &67.1\\
    DaP (seq) \cite{lee2021dialogue} & T5 Base (220M) & - & 51.2 & - & -\\
    DaP (ind) \cite{lee2021dialogue} & T5 Base (220M) & 56.7 & 57.6 & - & -\\
    \hline
    D3ST (Base) & T5 Base (220M) & 54.2 &56.1 &59.1 &72.1\\
    D3ST (Large) &  T5 Large (770M) & 54.5 &54.2 &58.6 &70.8\\
    D3ST (XXL) & T5 XXL (11B) & \textbf{57.8} &\textbf{58.7} &60.8 &\textbf{75.9}\\
    \hline
    \end{tabular}
    }
\caption{JGA on MultiWOZ 2.1-2.4.}
\end{subtable}

\vspace{0.5cm}

\begin{subtable}[ht]{\textwidth}
    \centering
    \scalebox{0.85}{
    \begin{tabular}{lcccc}
    \hline
    \textbf{Model} & \textbf{Pretrain. Model (\# Params.)} & \textbf{JGA} & \textbf{Intent} & \textbf{Req slot} \\
    \hline
    SGD baseline \cite{Rastogi_Zang_Sunkara_Gupta_Khaitan_2020} & BERT Base (110M) & 25.4 & 90.6 & 96.5 \\
    DaP (ind) \cite{lee2021dialogue} & T5 Base (220M) &  71.8 & 90.2 & 97.8\\
    SGP-DST \cite{ruan2020finetuning} & T5 Base (220M) & 72.2 & 91.8 & 99.0\\
    paDST\ding{110} \cite{ma2020endtoend} & XLNet Large (340M) & \textbf{86.5} & 94.8 & 98.5\\
    \hline
    D3ST (Base) & T5 Base (220M)  & 72.9 & 97.2 & 98.9\\
    D3ST (Large) & T5 Large (770M) & 80.0 & 97.1 & 99.1 \\
    D3ST (XXL) & T5 XXL (11B) & 86.4 & \textbf{98.8} & \textbf{99.4}  \\
    \hline
    \end{tabular}
    }
\caption{JGA, active intent accuracy and requested slot F1 on SGD.}
\end{subtable}
\caption{Results on MultiWOZ and SGD datasets with full training data.  ``-'' indicates no public number is available. Best results are marked in bold. \ding{72}: SimpleTOD results are retrieved from the 2.3 website \url{https://github.com/lexmen318/MultiWOZ-coref}, in which two numbers are reported for 2.1 (one produced by the 2.3 author, the other by the original SimpleTOD paper). \ding{70}: No data pre-processing applied for MultiWOZ 2.1. \ding{110}: Data augmentation and special rules applied.}
\label{table:full_comparison}
\end{center}
\end{table*}

Table \ref{table:full_comparison} gives the model quality when the entire training datasets are used for fine-tuning. We show that D3ST is close to, or at the state-of-the-art across all benchmarks, illustrating the effectiveness of the proposed approach. We also see that increasing the model size significantly improves the quality.

Note however that not all results are directly comparable, and we discuss some notable incongruities. The best result on SGD is from paDST, but this model has signicant advantages. paDST uses a data augmentation procedure by back-translating between English and Chinese, as well as special handcrafted rules for model predictions. In contrast, our models only train on the default SGD dataset, and do not apply any handcrafted rules whatsoever. While paDST has significantly higher JGA compared to the similarly-sized D3ST Large. D3ST XXL is on par, making up for its lack of data augmentation and handcrafted rules with a much larger model.

One other notable comparison can be made. DaP also relies on slot descriptions and is finetuned from a T5 Base model, making it directly comparable to our D3ST Base model, which exhibits better performance on SGD and MultiWOZ. One additional advantage of D3ST is that it predicts all slots at once in a single inference pass. In contrast, the independent (ind) decoding variant of DaP does inference once for every slot, similar to most other baselines, and is thus far less efficient. This is not scalable in TOD, especially with schemata becoming increasingly large in terms of the number of slots, intents, and domains.%

\subsection{Comparison of Description Types}
We now study whether the quality of D3ST is sensitive to the schema description types. For this, we run the same experiment as in Section \ref{sec:main_result} with D3ST Large and XXL, but using three different types of descriptions: human-readable language descriptions, schema element names (abbreviations) as defined in the original schema, and random strings. The random string descriptions are generated by simply randomly permuting the character sequences of the original element names. This experiment is designed to check how a model with only memorization capability without any understanding of schema element semantics does on seen and unseen schemas. An example of all three description type comparisons can be found in Appendix \ref{sec:example_of_desc_types}.

\begin{table}[htbp]
\small
\centering
    \centering
    \scalebox{0.95}{
    \begin{tabular}{lccccc}
    \hline
    \textbf{Type} & \textbf{M2.1} & \textbf{M2.2} & \textbf{M2.3} & \textbf{M2.4} & \textbf{SGD} \\
    \hline
    \multirow{2}{*}{Language} &54.5 &55.9 &58.6 &70.8 &80.0 \\&57.8 &58.7 &60.8 &75.9 &86.4 \\
    \hline
    \multirow{2}{*}{Name} &55.1 &55.8 &59.6 &72.2 &73.7 \\&57.5 &57.9 &60.4 &75.4 &79.7 \\
    \hline
    \multirow{2}{*}{Random} &20.1 &9.0 &12.1 &16.9 &37.4
\\ &57.6 &56.1 &59.3 &73.6 &64.8 \\
    \hline
    \end{tabular}
    }
    \caption{Comparison between D3ST models using different types of descriptions on MultiWOZ and SGD. ``Language'', ``Name'' and ``Random'' correspond to using detailed language description, schema element name and random strings respectively. Each type contains two rows, corresponding to the results given by ``large'' and ``XXL'' models. Note that the "Random" experiments for "large" models had trouble converging, and we instead report their JGA at 85k steps.}
    \label{table:description_comparison}
\end{table}

Table \ref{table:description_comparison} compares the performance with different description types. It can be seen that using language descriptions consistently outperforms other types, aligned with our expectation that natural and human-readable descriptions contain richer semantics and are aligned with the pretraining objective, enabling LM to perform better. Element names are less readable than full descriptions, but still retain some semantics: they preform well but fall short of full descriptions. On the other hand, using random strings performs worst on average, even on MultiWOZ where the training and test schema are the same (and the model is allowed to memorize descriptions from training). With random strings, there is the extra challenge of identifying the correct slot id for each value to predict, since each example has a random shuffling of the slot ids. Indeed, we observed that training "large" models on random names is hard to converge, and instead of reporting their final results, we stopped these experiments early and reported their JGA at 85k steps. The XXL models did not encounter the same issue; we suspect that it was easier for larger models to memorize slot name permutations.

In constrast to MultiWOZ, SGD requires models to generalize to unseen tasks and domains in the evaluation datasets. Here, using random strings undermined quality significantly. 
In general, meaningless inputs hurt performance and lead to less generalization. We therefore suggest instructing the model with semantically rich representations, in particular, language descriptions.

One more observation we make is that, on large MultiWOZ models, using element names had better JGA than using a full language description. This trend does not hold on SGD, and also reverses when trained with XXL. We hypothesize that this is a result of input sequence length: on MultiWOZ we feed slots descriptions from all domains as prefix, and when full language description is utilized, the input sequence becomes excessively long. Using element names shortens the length, making a moderate-size model easier to learn. In contrast, input sequence lengths on SGD are lower than that on MultiWOZ, since only active domains are provided as part of the input.

\subsection{Data Efficiency}
Properly designed prefixes or prompts have been shown to significantly improve an LM's data efficiency \citep{radford2019language,liu2021pretrain,wei2021finetuned}. We investigate how different types of description prefixes vary in performance in low-resource regimes by running experiments with large and XXL models on SGD with 0.16\% (10-shot), 1\%, and 10\% of training data. For the 0.16\% experiment, we randomly select 10 samples from each training domain to increase the domain diversity, totalling 260 examples. For other experiments the samples are uniformly sampled across the entire training set. 
We sample from three random seeds for each experiment.

\begin{table}[htbp]
\centering
    \centering
    \scalebox{0.9}{
    \begin{tabular}{lccc}
    \hline
    \textbf{Type} & \textbf{0.18\%} & \textbf{1\%} & \textbf{10\%} \\
    \hline
    \multirow{2}{*}{Language} &6.1 $\pm$ 0.7 &36.7 $\pm$ 2.0 &73.1 $\pm$ 0.2 \\&51.0 $\pm$ 0.2 &79.4 $\pm$ 0.4 &83.0 $\pm$ 0.1 \\
    \hline
    \multirow{2}{*}{Name} &5.0 $\pm$ 0.2 &28.0 $\pm$ 2.7 &69.7 $\pm$ 0.3 \\&47.7 $\pm$ 0.5 &74.9 $\pm$ 1.4 &78.6 $\pm$ 0.7 \\
    \hline
    \end{tabular}
    }
    \caption{Data efficiency of D3ST using natural language and element name descriptions, trained and evaluated on SGD. Each description type contains two rows, corresponding to the results given by ``large'' and ``XXL'' models. The metric is JGA.}
    \label{table:description_data_efficiency}
\end{table}

The results are given in Table \ref{table:description_data_efficiency}. From the table we have the following observations:
\begin{itemize}
    \item Using human-readable language descriptions consistently outperforms other types of representations, indicating better data efficiency with semantically-rich descriptions.
    \item With just 0.18\% of the data, XXL models can already reach more than half of their full quality (from Table \ref{table:full_comparison}). At 1\%, we observe quality close to using 100\% data. Increasing to 10\% only yielded marginal gains.
    \item Larger models are much more data efficient than smaller ones, as can be seen from the big gap between ``large'' and ``XXL'' models.
\end{itemize}

\subsection{Zero-shot Transfer to Unseen Tasks}

To assess our approach's zero-shot transfer ability to unseen tasks, we conduct the following set of experiments:

\noindent\textbf{MultiWOZ cross-domain transfer} Following a setup similar to TransferQA \citep{lin2021zeroshot} and T5DST \citep{lin-etal-2021-leveraging}, we run the ``leave-one-out'' cross-domain zero-shot transfer evaluation on MultiWOZ 2.1.\footnote{For zero-shot evaluation, \citet{lin2021zeroshot} and \citet{lin-etal-2021-leveraging} experimented on MultiWOZ 2.1 and 2.0 respectively. While our models are trained and evaluated on MultiWOZ 2.1, we include results from both of them for comparison.} For each domain, we train a model on examples excluding that domain, and evaluate it on examples including it. Table \ref{table:cross_domain_mw} shows our results in comparison with the baselines.\footnote{When skipping the \texttt{train} domain, we postprocess predictions for slots \texttt{train-departure} and \texttt{train-destination} by ignoring the suffix "train station". This is semantically correct and improves JGA.} It can be seen that our approach achieves the best cross-domain transfer performance with significant gains across almost all domains.

\noindent\textbf{SGD unseen service transfer} The SGD benchmark contains numerous services and some domains only present in the test set. We present the results for zero-shot transfer to these domains and services in Table \ref{table:seen_unseen_sgd}. Note that D3ST Base has worse JGA on unseen domains when fairly compared to DaP and SGP-DST. However, D3ST has superlative JGA on seen domains, even better than paDST (with data augmentation and hand-crafted rules). In addition, increasing the size of D3ST further increases both seen and especially unseen JGA, indicating better generalization. At XXL, JGA on unseen domains is almost equal to paDST. %
 
\noindent\textbf{Cross-dataset transfer} In this setup, we evaluate if a model trained on one dataset can be directly applied to another dataset. To this end,
we train a model on SGD then directly evaluate on the MultiWOZ 2.4 test set, and vice versa\footnote{Note that the SGD dataset defines the services that will occur in each dialogue, whereas MultiWOZ expects models to be able to predict any of its domains for all dialogues. 
To make it compatible between SGD and MultiWOZ for cross-task zero-shot transfer, we limit the schema prefix for MutliWOZ to domains that appear in the current dialogue.
}. In both cases we use the XXL model from Section \ref{sec:main_result}, and report the numbers in Table \ref{table:zero_shot}.

Despite obvious schema differences and domain mismatch between MultiWOZ and SGD, our model trained on MultiWOZ already achieves zero-shot quality on SGD close to the BERT-baseline \citep{Rastogi_Zang_Sunkara_Gupta_Khaitan_2020} with 25.4\% JGA. Our model trained on SGD and evaluated on MultiWOZ shows similarly strong zero-shot results. Both results are much lower than the state of the art for both datasets however, due to differing biases defined in schemata between the two datasets, and from latent knowledge that isn't captured from a schema alone.

\begin{table}[htbp]
\small
\begin{subtable}[ht]{0.45\textwidth}
\centering
    \centering
    \scalebox{0.95}{
    \begin{tabular}{lccc}
    \hline
    \multirow{2}{4em}{\textbf{Domain}} & \multicolumn{2}{c}{\textbf{JGA}}\\
    & D3ST & TransferQA & T5DST \\
    \hline
    Attraction &\textbf{56.4} & 31.3 &33.1 \\
    Hotel &21.8 & \textbf{22.7} &21.2 \\
    Restaurant &\textbf{38.2} & 26.3 &21.7 \\
    Taxi &\textbf{78.4} & 61.9 &64.6 \\
    Train &\textbf{38.7} & 36.7 &35.4 \\
    \hline
    Avg &\textbf{46.7} & 35.8 & 35.2 \\
    \hline
    \end{tabular}
    }
\caption{Cross-domain (leave-one-out) transfer on MultiWOZ.}
\label{table:cross_domain_mw}
\end{subtable}

\vspace{0.2cm}

\begin{subtable}[ht]{0.5\textwidth}
\centering
    \centering
    \scalebox{0.95}{
    \begin{tabular}{lccc}
    \hline
    \multirow{2}{4em}{Model} & \multicolumn{3}{c}{\textbf{JGA}}\\
    & Overall & Seen & Unseen\\
    \hline
    SGD Baseline & 25.4 & 41.2 & 20.0 \\
    DaP (ind) & 71.8 & 83.3 & 68.0 \\
    SGP-DST & 72.2 & 87.9 & 66.9 \\
    Team14\ding{115} & 77.3 & 90.0 & 73.0 \\
    paDST\ding{110} & \textbf{86.5} &92.4 & \textbf{84.6} \\
    \hline
    D3ST (base) & 72.9 & 92.5 & 66.4\\
    D3ST (large) & 80.0 & 93.8 & 75.4\\
    D3ST (XXL) & 86.4 & \textbf{95.8} & 83.3\\
    \hline
    \end{tabular}
    }
\caption{JGA on seen versus unseen services for SGD. \ding{115} and \ding{110} have the same meaning as in Table \ref{table:full_comparison}.}
\label{table:seen_unseen_sgd}
\end{subtable}

\vspace{0.2cm}

\begin{subtable}[ht]{0.5\textwidth}
\centering
    \centering
    \scalebox{0.95}{
    \begin{tabular}{lc}
    \hline
    \textbf{Transfer} & \textbf{JGA} \\
    \hline
    SGD$\rightarrow$MultiWOZ & 28.9 \\
    MultiWOZ$\rightarrow$SGD & 23.1 \\
    \hline
    \end{tabular}
    }
\caption{Cross-dataset transfer b/w SGD and MultiWOZ 2.4.}
\end{subtable}
\caption{Zero-shot transfer evaluation results from three different setups.}
\label{table:zero_shot}
\end{table}

\noindent\textbf{Qualitative Evaluation} In addition to quantitatively evaluating zero-shot transfer, we qualitatively examined examples of D3ST transferring to novel domains. We handcrafted a few dialogues for domains very different from the ones seen in the SGD dataset (e.g. conference submission, internet provider, e-commerce retailer). We designed the dialogues to be as stylistically realistic as possible for customer service scenarios. We tasked the XXL model trained on SGD (from Table \ref{table:full_comparison}) with inferring their dialogue states, and share one example in Table \ref{table:zero_shot_example}. More examples can be found in Table \ref{tab:zero_shot_example_appx} of Appendix \ref{sec:zero_shot_example_appx}. We observe that the model performs surprisingly well across all of our handcrafted dialogues, even though the domains are very different from the training data.

\begin{table}[htbp]
\small
\centering
    \centering
    \scalebox{0.8}{
    \begin{tabularx}{\columnwidth}{|X|}
    \hline
    \textbf{Domain: } Conference Submission\\
    \\
    \textbf{Input:} \texttt{\textcolor{red}{0:name of the conference 1:title of the paper 2:the first author of the paper 3:research areas for the paper 4:email for openreview account i1:submit a paper to a conference i2:check if a paper has been accepted} \textcolor{blue}{[user] hi, i'd like to submit a paper for a conference [system] that's great. which conference would you like to submit to? [user] i'd like to submit to \textbf{acl 2022} [system] ok. could you share the title of your paper and the name of your first author? [user] the paper is "\textbf{description-driven task-oriented dialog modeling}", and the first author is \textbf{grace hopper} [system] great, thank you. note that this year, we require all paper authors to be registered on openreview. could you give the email for your openreview account? [user] sure, its \textbf{gracehopper@gmail.com}}} \\
    \\
    \textbf{Prediction:}\texttt{\textcolor{teal}{ [states] 0:acl 2022 1:description-driven task-oriented dialog modeling 2:grace hopper 4:gracehopper@gmail.com [intents] i1}} \\
    \hline
    \end{tabularx}
    }
    \caption{An example of D3ST performing zero-shot transfer to a hypothetical "Conference Submission" domain. The predicted dialogue state is entirely correct. Boldface and color were added for visual clarity.}
    \label{table:zero_shot_example}
\end{table}

\subsection{Robustness to Variations of Descriptions}
Since there are many ways to provide descriptions for a given schema, a natural question to raise about this approach is how robust the model is against different choices of descriptions. The recently proposed SGD-X benchmark \citep{lee2021sgdx} is designed specifically for the study of this problem. SGD-X contains five variations of the original SGD, each one using a different set of schema descriptions provided by different crowd-source workers. To assess the robustness of D3ST, we use the large and XXL models evaluated in Section~\ref{sec:main_result} and decode test sets from each of the five variants of SGD-X. A robust model is expected to have smaller fluctuations in predictions across schema variants for the same dialogue context, as measured by Schema Sensitivity \texttt{SS(JGA)} defined in \citet{lee2021sgdx},. which calculates the average variation coefficient of JGA at turn level. A lower \texttt{SS(JGA)} value implies less fluctuation and more robustness.

We compare the robustness of models using different prompt types in Table \ref{table:robustness}. From the numbers we see that using the most human-readable natural language descriptions not only achieves the highest average accuracy over all SGD-X test set variants, but also enjoys the smallest \texttt{SS(JGA)} at the same model size. This indicates that description-driven models are more robust. On the other hand, using element names and random names have progressively lower mean accuracy and higher sensitivity to schema changes.

\begin{table}[htbp]
\begin{subtable}[ht]{0.45\textwidth}
\small
\setlength{\tabcolsep}{3pt}
\centering
    \centering
    \scalebox{0.9}{
    \begin{tabular}{lcccccccc}
    \hline
    \textbf{Size} & \textbf{Orig} & \textbf{v1} & \textbf{v2} & \textbf{v3} & \textbf{v4} & \textbf{v5} & \textbf{Avg v1-5} & \textbf{SS(JGA)}\\
    \hline
    large &80.0 & 79.9 & 79.4	&76.5 &71.9 &69.1 & 75.3 & 0.26\\
    XXL &86.4 & 85.5 &85.1 &73.9 &75.5 &68.9 & 77.8 & 0.27\\
    \hline
    \end{tabular}
    }
\caption{Natural language description}
\end{subtable}

\begin{subtable}[ht]{0.45\textwidth}
\small
\setlength{\tabcolsep}{3pt}
\centering
    \centering
    \scalebox{0.9}{
    \begin{tabular}{lcccccccc}
    \hline
    \textbf{Size} & \textbf{Orig} & \textbf{v1} & \textbf{v2} & \textbf{v3} & \textbf{v4} & \textbf{v5} & \textbf{Avg v1-5} & \textbf{SS(JGA)}\\
    \hline
    large &73.7 & 72	&69.5 &66.4	&61.1 &65.7 & 66.9 &0.37\\
    XXL &79.7 & 80.8 &76.6 &74.2 &61.2 &72.3 & 73.0 &0.35\\
    \hline
    \end{tabular}
    }
\caption{Element name description}    
\end{subtable}

\begin{subtable}[ht]{0.45\textwidth}
\small
\setlength{\tabcolsep}{3pt}
\centering
    \centering
    \scalebox{0.9}{
    \begin{tabular}{lcccccccc}
    \hline
    \textbf{Size} & \textbf{Orig} & \textbf{v1} & \textbf{v2} & \textbf{v3} & \textbf{v4} & \textbf{v5} & \textbf{Avg v1-5} & \textbf{SS(JGA)}\\
    \hline
    large &37.4 &29.3&	34.6&	28.0&	25.2&	25.0 & 28.4 & 0.74 \\
    XXL &64.8 &67.8 &	68.8&	72.9	&58.1&	68.1&	67.1 & 0.51\\
    \hline
    \end{tabular}
    }
\caption{Random description}
\end{subtable}
\caption{Robustness comparison for various description types. SS(JGA) refers to schema sensitivity for JGA.}
\label{table:robustness}
\end{table}

\section{Conclusion}
We advocate using human-readable language descriptions in place of abbreviated or arbitrary notations for schema definition in TOD modeling. We believe this schema representation contains more meaningful information for a strong LM to leverage, leading to better performance and improved data efficiency. To this end, we propose a simple and effective DST model named ``Description-Driven Dialogue State Tracking'' (D3ST), which relies fully on schema descriptions and an index-picking mechanism to indicate active slots or intents. Our experiments verify the effectiveness of description-driven dialogue modeling in the following ways. First, D3ST achieves superior quality on MultiWOZ and SGD. Second, using language descriptions outperforms abbreviations or arbitrary notations. Third, the description driven approach improves data-efficiency, and enables effective zero-shot transfer to unseen tasks and domains. Fourth, using language for schema description improves model robustness as measured by the SGD-X benchmark.

\bibliography{anthology,custom}

\begin{thebibliography}{35}
\expandafter\ifx\csname natexlab\endcsname\relax\def\natexlab#1{#1}\fi

\bibitem[{Baolin~Peng(2020)}]{peng2020scgpt}
Chunyuan Li Xiujun Li Jinchao Li Michael Zeng Jianfeng~Gao Baolin~Peng,
  Chenguang~Zhu. 2020.
\newblock \href {http://arxiv.org/abs/2002.12328} {Few-shot natural language
  generation for task-oriented dialog}.

\bibitem[{Brown et~al.(2020)Brown, Mann, Ryder, Subbiah, Kaplan, Dhariwal,
  Neelakantan, Shyam, Sastry, Askell, Agarwal, Herbert-Voss, Krueger, Henighan,
  Child, Ramesh, Ziegler, Wu, Winter, Hesse, Chen, Sigler, Litwin, Gray, Chess,
  Clark, Berner, McCandlish, Radford, Sutskever, and
  Amodei}]{NEURIPS2020_1457c0d6}
Tom Brown, Benjamin Mann, Nick Ryder, Melanie Subbiah, Jared~D Kaplan, Prafulla
  Dhariwal, Arvind Neelakantan, Pranav Shyam, Girish Sastry, Amanda Askell,
  Sandhini Agarwal, Ariel Herbert-Voss, Gretchen Krueger, Tom Henighan, Rewon
  Child, Aditya Ramesh, Daniel Ziegler, Jeffrey Wu, Clemens Winter, Chris
  Hesse, Mark Chen, Eric Sigler, Mateusz Litwin, Scott Gray, Benjamin Chess,
  Jack Clark, Christopher Berner, Sam McCandlish, Alec Radford, Ilya Sutskever,
  and Dario Amodei. 2020.
\newblock \href
  {https://proceedings.neurips.cc/paper/2020/file/1457c0d6bfcb4967418bfb8ac142f64a-Paper.pdf}
  {Language models are few-shot learners}.
\newblock In \emph{Advances in Neural Information Processing Systems},
  volume~33, pages 1877--1901. Curran Associates, Inc.

\bibitem[{Budzianowski and Vuli{\'c}(2019)}]{budzianowski-vulic-2019-hello}
Pawe{\l} Budzianowski and Ivan Vuli{\'c}. 2019.
\newblock \href {https://doi.org/10.18653/v1/D19-5602} {Hello, it{'}s {GPT}-2 -
  how can {I} help you? towards the use of pretrained language models for
  task-oriented dialogue systems}.
\newblock In \emph{Proceedings of the 3rd Workshop on Neural Generation and
  Translation}, pages 15--22, Hong Kong. Association for Computational
  Linguistics.

\bibitem[{Budzianowski et~al.(2018)Budzianowski, Wen, Tseng, Casanueva, Ultes,
  Ramadan, and Ga{\v{s}}i{\'c}}]{budzianowski-etal-2018-multiwoz}
Pawe{\l} Budzianowski, Tsung-Hsien Wen, Bo-Hsiang Tseng, I{\~n}igo Casanueva,
  Stefan Ultes, Osman Ramadan, and Milica Ga{\v{s}}i{\'c}. 2018.
\newblock \href {https://doi.org/10.18653/v1/D18-1547} {{M}ulti{WOZ} - a
  large-scale multi-domain {W}izard-of-{O}z dataset for task-oriented dialogue
  modelling}.
\newblock In \emph{Proceedings of the 2018 Conference on Empirical Methods in
  Natural Language Processing}, pages 5016--5026, Brussels, Belgium.
  Association for Computational Linguistics.

\bibitem[{Du et~al.(2021)Du, He, Li, Yu, Pasupat, and Zhang}]{du-etal-2021-qa}
Xinya Du, Luheng He, Qi~Li, Dian Yu, Panupong Pasupat, and Yuan Zhang. 2021.
\newblock \href {https://doi.org/10.18653/v1/2021.acl-short.83} {{QA}-driven
  zero-shot slot filling with weak supervision pretraining}.
\newblock In \emph{Proceedings of the 59th Annual Meeting of the Association
  for Computational Linguistics and the 11th International Joint Conference on
  Natural Language Processing (Volume 2: Short Papers)}, pages 654--664,
  Online. Association for Computational Linguistics.

\bibitem[{Gao et~al.(2020)Gao, Agarwal, Jin, Chung, and
  Hakkani-Tur}]{gao-etal-2020-machine}
Shuyang Gao, Sanchit Agarwal, Di~Jin, Tagyoung Chung, and Dilek Hakkani-Tur.
  2020.
\newblock \href {https://doi.org/10.18653/v1/2020.nlp4convai-1.10} {From
  machine reading comprehension to dialogue state tracking: Bridging the gap}.
\newblock In \emph{Proceedings of the 2nd Workshop on Natural Language
  Processing for Conversational AI}, pages 79--89, Online. Association for
  Computational Linguistics.

\bibitem[{Han et~al.(2021)Han, Liu, Takanobu, Lian, Huang, Wan, Peng, and
  Huang}]{han2021multiwoz}
Ting Han, Ximing Liu, Ryuichi Takanobu, Yixin Lian, Chongxuan Huang, Dazhen
  Wan, Wei Peng, and Minlie Huang. 2021.
\newblock \href {http://arxiv.org/abs/2010.05594} {Multiwoz 2.3: A multi-domain
  task-oriented dialogue dataset enhanced with annotation corrections and
  co-reference annotation}.

\bibitem[{Heck et~al.(2020)Heck, van Niekerk, Lubis, Geishauser, Lin, Moresi,
  and Gasic}]{heck-etal-2020-trippy}
Michael Heck, Carel van Niekerk, Nurul Lubis, Christian Geishauser, Hsien-Chin
  Lin, Marco Moresi, and Milica Gasic. 2020.
\newblock \href {https://aclanthology.org/2020.sigdial-1.4} {{T}rip{P}y: A
  triple copy strategy for value independent neural dialog state tracking}.
\newblock In \emph{Proceedings of the 21th Annual Meeting of the Special
  Interest Group on Discourse and Dialogue}, pages 35--44, 1st virtual meeting.
  Association for Computational Linguistics.

\bibitem[{Hosseini-Asl et~al.(2020)Hosseini-Asl, McCann, Wu, Yavuz, and
  Socher}]{hosseiniasl2020simple}
Ehsan Hosseini-Asl, Bryan McCann, Chien-Sheng Wu, Semih Yavuz, and Richard
  Socher. 2020.
\newblock \href {http://arxiv.org/abs/2005.00796} {A simple language model for
  task-oriented dialogue}.

\bibitem[{Jouppi et~al.(2017)Jouppi, Young, Patil, Patterson, Agrawal, Bajwa,
  Bates, Bhatia, Boden, Borchers, Boyle, and Cantin}]{tpu2017}
Norman~P. Jouppi, Cliff Young, Nishant Patil, David Patterson, Gaurav Agrawal,
  Raminder Bajwa, Sarah Bates, Suresh Bhatia, Nan Boden, Al~Borchers, Rick
  Boyle, and Pierre-luc et~al. Cantin. 2017.
\newblock \href {https://doi.org/10.1145/3140659.3080246} {In-datacenter
  performance analysis of a tensor processing unit}.
\newblock \emph{SIGARCH Comput. Archit. News}, 45(2):1–12.

\bibitem[{Kim et~al.(2020)Kim, Yang, Kim, and Lee}]{kim2020efficient}
Sungdong Kim, Sohee Yang, Gyuwan Kim, and Sang-Woo Lee. 2020.
\newblock \href {http://arxiv.org/abs/1911.03906} {Efficient dialogue state
  tracking by selectively overwriting memory}.

\bibitem[{Lee et~al.(2021{\natexlab{a}})Lee, Cheng, and
  Ostendorf}]{lee2021dialogue}
Chia-Hsuan Lee, Hao Cheng, and Mari Ostendorf. 2021{\natexlab{a}}.
\newblock Dialogue state tracking with a language model using schema-driven
  prompting.
\newblock In \emph{Proceedings of the 2021 Conference on Empirical Methods in
  Natural Language Processing (EMNLP)}.

\bibitem[{Lee et~al.(2021{\natexlab{b}})Lee, Gupta, Rastogi, Cao, Zhang, and
  Wu}]{lee2021sgdx}
Harrison Lee, Raghav Gupta, Abhinav Rastogi, Yuan Cao, Bin Zhang, and Yonghui
  Wu. 2021{\natexlab{b}}.
\newblock \href {http://arxiv.org/abs/2110.06800} {Sgd-x: A benchmark for
  robust generalization in schema-guided dialogue systems}.

\bibitem[{Li et~al.(2021)Li, Cao, Sridhar, Zhu, Li, Hamza, and
  McAuley}]{li-etal-2021-zero}
Shuyang Li, Jin Cao, Mukund Sridhar, Henghui Zhu, Shang-Wen Li, Wael Hamza, and
  Julian McAuley. 2021.
\newblock \href {https://aclanthology.org/2021.eacl-main.91} {Zero-shot
  generalization in dialog state tracking through generative question
  answering}.
\newblock In \emph{Proceedings of the 16th Conference of the European Chapter
  of the Association for Computational Linguistics: Main Volume}, pages
  1063--1074, Online. Association for Computational Linguistics.

\bibitem[{Lin et~al.(2021{\natexlab{a}})Lin, Liu, Madotto, Moon, Crook, Zhou,
  Wang, Yu, Cho, Subba, and Fung}]{lin2021zeroshot}
Zhaojiang Lin, Bing Liu, Andrea Madotto, Seungwhan Moon, Paul Crook, Zhenpeng
  Zhou, Zhiguang Wang, Zhou Yu, Eunjoon Cho, Rajen Subba, and Pascale Fung.
  2021{\natexlab{a}}.
\newblock \href {http://arxiv.org/abs/2109.04655} {Zero-shot dialogue state
  tracking via cross-task transfer}.

\bibitem[{Lin et~al.(2021{\natexlab{b}})Lin, Liu, Moon, Crook, Zhou, Wang, Yu,
  Madotto, Cho, and Subba}]{lin-etal-2021-leveraging}
Zhaojiang Lin, Bing Liu, Seungwhan Moon, Paul Crook, Zhenpeng Zhou, Zhiguang
  Wang, Zhou Yu, Andrea Madotto, Eunjoon Cho, and Rajen Subba.
  2021{\natexlab{b}}.
\newblock \href {https://doi.org/10.18653/v1/2021.naacl-main.448} {Leveraging
  slot descriptions for zero-shot cross-domain dialogue {S}tate{T}racking}.
\newblock In \emph{Proceedings of the 2021 Conference of the North American
  Chapter of the Association for Computational Linguistics: Human Language
  Technologies}, pages 5640--5648, Online. Association for Computational
  Linguistics.

\bibitem[{Liu et~al.(2021)Liu, Yuan, Fu, Jiang, Hayashi, and
  Neubig}]{liu2021pretrain}
Pengfei Liu, Weizhe Yuan, Jinlan Fu, Zhengbao Jiang, Hiroaki Hayashi, and
  Graham Neubig. 2021.
\newblock \href {http://arxiv.org/abs/2107.13586} {Pre-train, prompt, and
  predict: A systematic survey of prompting methods in natural language
  processing}.

\bibitem[{Ma et~al.(2020)Ma, Zeng, Zhu, Li, Yang, Yao, Zhou, and
  Shen}]{ma2020endtoend}
Yue Ma, Zengfeng Zeng, Dawei Zhu, Xuan Li, Yiying Yang, Xiaoyuan Yao, Kaijie
  Zhou, and Jianping Shen. 2020.
\newblock \href {http://arxiv.org/abs/1912.09297} {An end-to-end dialogue state
  tracking system with machine reading comprehension and wide \& deep
  classification}.

\bibitem[{Madotto et~al.(2021)Madotto, Lin, Winata, and
  Fung}]{madotto2021fewshot}
Andrea Madotto, Zhaojiang Lin, Genta~Indra Winata, and Pascale Fung. 2021.
\newblock \href {http://arxiv.org/abs/2110.08118} {Few-shot bot: Prompt-based
  learning for dialogue systems}.

\bibitem[{Madotto et~al.(2020)Madotto, Liu, Lin, and
  Fung}]{madotto2020language}
Andrea Madotto, Zihan Liu, Zhaojiang Lin, and Pascale Fung. 2020.
\newblock \href {http://arxiv.org/abs/2008.06239} {Language models as few-shot
  learner for task-oriented dialogue systems}.

\bibitem[{Mi et~al.(2021)Mi, Li, Wang, Jiang, and Liu}]{mi2021cins}
Fei Mi, Yitong Li, Yasheng Wang, Xin Jiang, and Qun Liu. 2021.
\newblock \href {http://arxiv.org/abs/2109.04645} {Cins: Comprehensive
  instruction for few-shot learning in task-oriented dialog systems}.

\bibitem[{Mishra et~al.(2021)Mishra, Khashabi, Baral, and
  Hajishirzi}]{mishra2021crosstask}
Swaroop Mishra, Daniel Khashabi, Chitta Baral, and Hannaneh Hajishirzi. 2021.
\newblock \href {http://arxiv.org/abs/2104.08773} {Cross-task generalization
  via natural language crowdsourcing instructions}.

\bibitem[{Namazifar et~al.(2020)Namazifar, Papangelis, Tur, and
  Hakkani-Tür}]{namazifar2020language}
Mahdi Namazifar, Alexandros Papangelis, Gokhan Tur, and Dilek Hakkani-Tür.
  2020.
\newblock \href {http://arxiv.org/abs/2011.03023} {Language model is all you
  need: Natural language understanding as question answering}.

\bibitem[{Radford et~al.(2019)Radford, Wu, Child, Luan, Amodei, and
  Sutskever}]{radford2019language}
Alec Radford, Jeff Wu, Rewon Child, David Luan, Dario Amodei, and Ilya
  Sutskever. 2019.
\newblock Language models are unsupervised multitask learners.

\bibitem[{Raffel et~al.(2020)Raffel, Shazeer, Roberts, Lee, Narang, Matena,
  Zhou, Li, and Liu}]{2020t5}
Colin Raffel, Noam Shazeer, Adam Roberts, Katherine Lee, Sharan Narang, Michael
  Matena, Yanqi Zhou, Wei Li, and Peter~J. Liu. 2020.
\newblock \href {http://jmlr.org/papers/v21/20-074.html} {Exploring the limits
  of transfer learning with a unified text-to-text transformer}.
\newblock \emph{Journal of Machine Learning Research}, 21(140):1--67.

\bibitem[{Rastogi et~al.(2020)Rastogi, Zang, Sunkara, Gupta, and
  Khaitan}]{Rastogi_Zang_Sunkara_Gupta_Khaitan_2020}
Abhinav Rastogi, Xiaoxue Zang, Srinivas Sunkara, Raghav Gupta, and Pranav
  Khaitan. 2020.
\newblock \href {https://doi.org/10.1609/aaai.v34i05.6394} {Towards scalable
  multi-domain conversational agents: The schema-guided dialogue dataset}.
\newblock \emph{Proceedings of the AAAI Conference on Artificial Intelligence},
  34(05):8689--8696.

\bibitem[{Ruan et~al.(2020)Ruan, Ling, Gu, and Liu}]{ruan2020finetuning}
Yu-Ping Ruan, Zhen-Hua Ling, Jia-Chen Gu, and Quan Liu. 2020.
\newblock \href {http://arxiv.org/abs/2002.00181} {Fine-tuning bert for
  schema-guided zero-shot dialogue state tracking}.

\bibitem[{Shah et~al.(2019)Shah, Gupta, Fayazi, and
  Hakkani-Tur}]{shah-etal-2019-robust}
Darsh Shah, Raghav Gupta, Amir Fayazi, and Dilek Hakkani-Tur. 2019.
\newblock \href {https://doi.org/10.18653/v1/P19-1547} {Robust zero-shot
  cross-domain slot filling with example values}.
\newblock In \emph{Proceedings of the 57th Annual Meeting of the Association
  for Computational Linguistics}, pages 5484--5490, Florence, Italy.
  Association for Computational Linguistics.

\bibitem[{Wang et~al.(2020)Wang, Guo, and Zhu}]{wang-etal-2020-slot}
Yexiang Wang, Yi~Guo, and Siqi Zhu. 2020.
\newblock \href {https://doi.org/10.18653/v1/2020.emnlp-main.243} {Slot
  attention with value normalization for multi-domain dialogue state tracking}.
\newblock In \emph{Proceedings of the 2020 Conference on Empirical Methods in
  Natural Language Processing (EMNLP)}, pages 3019--3028, Online. Association
  for Computational Linguistics.

\bibitem[{Wei et~al.(2021)Wei, Bosma, Zhao, Guu, Yu, Lester, Du, Dai, and
  Le}]{wei2021finetuned}
Jason Wei, Maarten Bosma, Vincent~Y. Zhao, Kelvin Guu, Adams~Wei Yu, Brian
  Lester, Nan Du, Andrew~M. Dai, and Quoc~V. Le. 2021.
\newblock \href {http://arxiv.org/abs/2109.01652} {Finetuned language models
  are zero-shot learners}.

\bibitem[{Wu et~al.(2019)Wu, Madotto, Hosseini-Asl, Xiong, Socher, and
  Fung}]{wu-etal-2019-transferable}
Chien-Sheng Wu, Andrea Madotto, Ehsan Hosseini-Asl, Caiming Xiong, Richard
  Socher, and Pascale Fung. 2019.
\newblock \href {https://doi.org/10.18653/v1/P19-1078} {Transferable
  multi-domain state generator for task-oriented dialogue systems}.
\newblock In \emph{Proceedings of the 57th Annual Meeting of the Association
  for Computational Linguistics}, pages 808--819, Florence, Italy. Association
  for Computational Linguistics.

\bibitem[{Ye et~al.(2021)Ye, Manotumruksa, and Yilmaz}]{ye2021multiwoz}
Fanghua Ye, Jarana Manotumruksa, and Emine Yilmaz. 2021.
\newblock \href {http://arxiv.org/abs/2104.00773} {Multiwoz 2.4: A multi-domain
  task-oriented dialogue dataset with essential annotation corrections to
  improve state tracking evaluation}.

\bibitem[{Zang et~al.(2020)Zang, Rastogi, Sunkara, Gupta, Zhang, and
  Chen}]{zang-etal-2020-multiwoz}
Xiaoxue Zang, Abhinav Rastogi, Srinivas Sunkara, Raghav Gupta, Jianguo Zhang,
  and Jindong Chen. 2020.
\newblock \href {https://doi.org/10.18653/v1/2020.nlp4convai-1.13}
  {{M}ulti{WOZ} 2.2 : A dialogue dataset with additional annotation corrections
  and state tracking baselines}.
\newblock In \emph{Proceedings of the 2nd Workshop on Natural Language
  Processing for Conversational AI}, pages 109--117, Online. Association for
  Computational Linguistics.

\bibitem[{Zeng and Nie(2021)}]{zeng2021jointly}
Yan Zeng and Jian-Yun Nie. 2021.
\newblock \href {http://arxiv.org/abs/2010.14061} {Jointly optimizing state
  operation prediction and value generation for dialogue state tracking}.

\bibitem[{Zhao et~al.(2021)Zhao, Mahdieh, Zhang, Cao, and
  Wu}]{zhao-etal-2021-effective-sequence}
Jeffrey Zhao, Mahdis Mahdieh, Ye~Zhang, Yuan Cao, and Yonghui Wu. 2021.
\newblock \href {https://aclanthology.org/2021.emnlp-main.593} {Effective
  sequence-to-sequence dialogue state tracking}.
\newblock In \emph{Proceedings of the 2021 Conference on Empirical Methods in
  Natural Language Processing}, pages 7486--7493, Online and Punta Cana,
  Dominican Republic. Association for Computational Linguistics.

\end{thebibliography}
\bibliographystyle{acl_natbib}

\appendix
\setcounter{table}{0}
\renewcommand{\thetable}{A\arabic{table}}

\section{Example of Description Types}
\label{sec:example_of_desc_types}

An example of the different description types for a single example can be found in Table \ref{tab:example_of_desc_types_appx}.

\begin{table*}[ht]
\small
\centering
    \centering
    \begin{tabularx}{\linewidth}{r|X}
    Language & \texttt{\textcolor{red}{0:playback device on which the song is to be played 0a) bedroom speaker 0b) tv 0c) kitchen speaker 1=name of the artist the song is performed by 2=name of the song 3=album the song belongs to 4=genre of the song i0=search for a song based on the name and optionally other attributes i1=play a song by its name and optionally artist} \textcolor{blue}{[user] i want to find a movie. [system] what is your location. [user] santa rosa. i want to see it at 3rd street cinema. [system] i found 3 movies. does hellboy, how to train your dragon: the hidden world or the upside interest you? [user] how to train your dragon: the hidden world is perfect. can you find me some songs from the album summer anthems. [system] i found 1 song you may like. how about no other love from the album summer anthems by common kings? [user] that would be great. [system] play the song now? [user] play it on the bedroom device.}} \\
    \\
    Name & \texttt{\textcolor{red}{0:music\_2-genre 1:music\_2-playback\_device 1a) bedroom speaker 1b) kitchen speaker 1c) tv 2:music\_2-album 3:music\_2-artist 4:music\_2-song\_name i0:music\_2-playmedia i1:music\_2-lookupmusic} \textcolor{blue}{[user] i want to find a movie. [system] what is your location. [user] santa rosa. i want to see it at 3rd street cinema. [system] i found 3 movies. does hellboy, how to train your dragon: the hidden world or the upside interest you? [user] how to train your dragon: the hidden world is perfect. can you find me some songs from the album summer anthems. [system] i found 1 song you may like. how about no other love from the album summer anthems by common kings? [user] that would be great. [system] play the song now? [user] play it on the bedroom device.}} \\
    \\
    Random & \texttt{\textcolor{red}{0:e-e\_ciugs2mrn 1:psuekc\_l-2imceyibaca\_dv 1a) bedroom speaker 1b) kitchen speaker 1c) tv 2:umm2uisc\_bal- 3:satriti\_2-sumc 4:\_-onassng2\_cemmui i0:aeusmmci2-adipl\_y i1:miiu\_2olosckucp-ums} \textcolor{blue}{[user] i want to find a movie. [system] what is your location. [user] santa rosa. i want to see it at 3rd street cinema. [system] i found 3 movies. does hellboy, how to train your dragon: the hidden world or the upside interest you? [user] how to train your dragon: the hidden world is perfect. can you find me some songs from the album summer anthems. [system] i found 1 song you may like. how about no other love from the album summer anthems by common kings? [user] that would be great. [system] play the song now? [user] play it on the bedroom device.}} \\
    \\
    States & \texttt{\textcolor{teal}{[states] 1:1a 2:summer anthems 4:no other love [intents] i0}} \\
    \end{tabularx}
    \caption{Examples of the same SGD dialogue with different description types. "Language" uses a detailed natural language description, "Name" uses the schema element name, and "Random" is generated from a random shuffling of the slot name. Note that the categorical slot value enumeration is unaffected in "Random", and that all three description types would have the same target slots and intents.
    }
    \label{tab:example_of_desc_types_appx}
\end{table*}

\section{Zero-shot Transfer to Novel Domains}
\label{sec:zero_shot_example_appx}

Qualitative examples showcasing zero-shot transfer to novel domains can be found in Table \ref{tab:zero_shot_example_appx}.

\begin{table*}[ht]
\small
\centering
    \centering
    \begin{tabularx}{\linewidth}{r|X}
    Domain & Internet Provider \\
    \\
    Inputs & \texttt{\textcolor{red}{0:email address of the account 1:whether professional help is needed for internet installation 1a) true 1b) false 2:whether to bundle services on the same plan 2a) true 2b) false 3:download speed of the internet plan 4:whether services are for residential or business use 4a) residential 4b) business 5:the address to provide services to i0=buy or change an internet plan i1:file a formal complaint} \textcolor{blue}{[user] hi there - my internet contract is up for renewal, and i'm interested in exploring other plan options. [system] happy to help. is this for your home or for a business? [user] home [system] what's the email associated with your account? [user] noamchomsky@hotmail.com [system] thanks. your current plan is 25 mbps download speed for \$53 / month. the two other plans are 50 mbps for \$63 / month and 100 mbps for \$73 / month. would you interested in either of those? [user] i'm interested in upgrading to the 50 mbps plan. [system] great. for \$10 / month more, would you like to include our basic cable plan? [user] no thanks. i'll need to talk this over with my partner. thank you for your help.}} \\
    \\
    States & \texttt{\textcolor{teal}{[states] 0:noamchomsky@hotmail.com 3:50 mbps 4:4a [intents] i0}} \\
    \hline
    Domain & E-Commerce Retailer \\
    \\
    Inputs & \texttt{\textcolor{red}{0:phone number associated with the customer's account 1:a coupon code to apply to the purchase 2:the reason for the product return 2a) accidental purchase 2b) malfunction 2c) preference 3:the retail product to purchase or to be returned 4:date the product was purchased 5:identifier associated with the purchase i0:return a product i1:purchase a product} \textcolor{blue}{ [system] hi how can i help you today? [user] hello - i recently purchased a glow in the dark ball that i'd like to return. [system] no problem. i'm happy to help. can you provide the order number or date of purchase please? [user] 1ozdl3v260lkq, and i purchased it last week on nov 1, 2021 [system] thanks. and what's the reason for the return? [user] the ball seems to be broken. it doesn't actually glow in the dark. [system] sorry to hear about that. we'll process the return and you should receive a refund within 10 business days. is there anything else i can do for you? [user] no, thanks for your help!}} \\
    \\
    States & \texttt{\textcolor{teal}{[states] 2:2b 3:glow in the dark ball 4:nov 1, 2021 5:1ozdl3v260lkq [intents] i0}} \\
    \end{tabularx}
    \caption{Two more examples of D3ST trained on SGD performing zero-shot transfer to novel domains. The only error is in the "Internet Provider" example, where the model misses that the slot for "whether to bundle services on the same plan" should be false. We hypothesize that "bundle" is industry jargon that the model fails to associate with the dialogue context.
    }
    \label{tab:zero_shot_example_appx}
\end{table*}

\end{document}